# Accurate Pedestrian Tracking in Urban Canyons: A Multi-Modal Fusion Approach

Shahar Dubiner, Peng Ren, and Roberto Manduchi
University of California, Santa Cruz

## Abstract

The contribution describes a pedestrian navigation approach designed to improve localization accuracy in urban environments where GNSS performance is degraded, a problem that is especially critical for blind or low-vision users who depend on precise guidance such as identifying the correct side of a street. To address GNSS limitations and the impracticality of camera-based visual positioning, the work proposes a particle filter–based fusion of GNSS and inertial data that incorporates spatial priors from maps, such as impassable buildings and unlikely walking areas, functioning as a probabilistic form of map matching. Inertial localization is provided by the RoNIN machine learning method, and fusion with GNSS is achieved by weighting particles based on their consistency with GNSS estimates and uncertainty. The system was evaluated on six challenging walking routes in downtown San Francisco using three metrics related to sidewalk correctness and localization error. Results show that the fused approach (GNSS+RoNIN+PF) significantly outperforms GNSS-only localization on most metrics, while inertial-only localization with particle filtering also surpasses GNSS alone for critical measures such as sidewalk assignment and across-street error.

## Introduction

Navigation apps for pedestrians can be very useful for anyone finding themselves in an unfamiliar urban area. The localization accuracy of these systems, however, can be poor in "urban canyons", where tall buildings can occlude view to GNSS satellites. Visual Positioning Services, which uses images taken by the phone's camera, can be used to refine the user's localization, but may not be convenient to use while walking, or if the user is keeping the phone in their pocket. Accurate localization is particularly important for blind or low vision walkers, who rely on the navigation app for wayfinding and are unable to use visual referencing to correct for system inaccuracies. For example, determining which side of a street the user is located on is critical in order to direct the user to e.g. cross the street when necessary.

One well-established method for increasing the accuracy of GNSS localization is to use complementary information from inertial sensors, typically through Kalman filter-based fusion. In this work, we experimented with a different information fusion approach, based on particle filtering. Particle filtering, a widely used technique in robotic navigation, allows one to leverage prior information about the layout of the environment to be traversed. In our case, we use two types of spatial priors to condition the estimated user path: areas with impenetrable buildings; and street surfaces (except for crosswalks), where the user is unlikely to be walking on. Intuitively, this method can be seen as a "soft", probabilistic version of map matching. The inertial navigation component of the system relies on a machine learning algorithm (RoNIN; Herath et al., 2020), which in previous work was shown to produce robust, world-referenced localization. Fusion between GNSS and inertial localization is obtained through a simple algorithm, originally proposed by Ren & Ke, (2010), which modifies the weights of the particles in the filter depending on their distance to the GNSS location and to its radius of uncertainty.

In order to test the proposed algorithms, we recorded data collected by an iPhone carried by an experimenter as they walked through six paths in downtown San Francisco. These paths were located in areas with poor GNSS reception; two of them were in a covered transit hub at the bottom of a building (the Salesforce Transit Center). These paths are representative of real-world situations, and highlight the challenges of pedestrian tracking based solely on GNSS signals. We defined three different metrics. The first metric, Correct Sidewalk Assignment, measured the proportion of time in which the user was localized at or in the vicinity of the correct sidewalk in the correct city block. This coarse level of spatial information is critical for a navigation system to give correct directions. The second metric, Euclidean Error, measures the distance between the actual and the estimated location of the walker at each time. The third metric, Along- and Across-Street Error, decomposes the Euclidean error into orthogonal components, respectively parallel and orthogonal to the direction of the nearby street. Note that GNSS localization is known to typically produce higher across-street errors than along-street. We used these metrics to compare GNSS-only localization with inertial-only localization (using the RoNIN algorithm followed by particle filtering, dubbed RoNIN+PF), and GNSS-inertial fusion (still using RoNIN and particle filtering: GNSS+RoNIN+PF). In all but one metric (Along-Street Error), GNSS+RoNIN+PF produced substantially better results than GNSS only. RoNIN+PF (which doesn't use GNSS localization) was shown to provide better values of Correct Sidewalk Assignment and Across-Street Error than GNSS alone.

# Related Work

Navigating the concrete canyons of urban downtowns presents unique challenges for location-based technologies. The degradation of GNSS data in urban canyons stems from several physical phenomena. Direct line-of-sight to satellites becomes limited as buildings obstruct large portions of the sky. Signals that do reach the receiver often arrive via reflection off building surfaces, creating multipath interference. Additionally, signals passing through building materials suffer attenuation, further reducing their reliability.

It is well known that, in urban canyons, GNSS-based positioning is more accurate in the "along-street" direction than "across-street". This is because tall buildings at each side of a street may occlude satellites in the across-street direction". Knowledge of which satellites are not visible (because of occluded line of sight), combined with knowledge of the height of nearby buildings, can be used to approximately measure the users' across-street location through "shadow matching" (Wang et al., 2015), possibly fused with 3D-Mapping-Aided GNSS Ranging (Adjrad & Groves, 2018). Later work (Weng et al., 2023) showed that it is possible to estimate on which side of the street a pedestrian is located simply by identifying (through SNR analysis) which satellites are not in the line of sight (causing their signal to be affected by multipath distortion, if received at all).

Additional sensors (in particular, inertial sensors) can be employed to support localization when GNSS signal is absent or of poor quality. Importantly, unlike GNSS, inertial sensing can only provide relative positioning. Traditionally, inertial measurement units (IMU) and GNSS data are fused using the Kalman filter, but this gives poor results in the case of non-Gaussian error distributions, or when GNSS signal is lost (Rehman et al., 2020). One approach to dealing with periods of lost GNSS data is to employ machine learning to predict the error of IMU positioning. For example, Aslinezhad et al. (2020) used a neural network to predict missing GNSS measurements in a robust Kalman Filter scheme. Fang et al. (2020) used an LSTM network that is continuously trained when GNSS data is available. When GNSS data is lost, the LSTM, fed by inertial data, is used to generate pseudo GNSS increments that, after accumulation, provide the error signal for correcting IMU positioning in the Kalman filter. Similarly, Kim et al. (2021) trained a network for pedestrian dead-reckoning (PDR) prediction using "ground truth" data from GNSS (when available). Zhi et al. (2022) used a convolutional neural network - long short-term memory (CNN–LSTM) to compensate for situations when GNSS signal is unavailable. Niu et al. (2025) computed a measure of consistency between multi-epoch GNSS pseudo-range observations and PDR-based localization, in order to remove GNSS faulty measurements before fusion with PDR. Xu et al. (2023) used particle filtering to fuse PDR and GNSS locations, using a model originally proposed by Ren & Ke (2010). Unlike our method, though, they do not leverage prior knowledge of the location of impenetrable buildings or of street surfaces. Estimation of the user's direction using PDR can also be used to remove unlikely sidewalks from search (Wend et al., 2024).

Map matching is commonly used for car positioning, using Kalman filtering or particle filtering from GNSS data combined with odometry from the vehicle (Quddus et al., 2007). Weng et al (2025) force the location of the pedestrian to be on a sidewalk, where the sidewalk identification (which side of the street the user is on) is determined by identifying the side of the sky from which most LOS signals are detected. A PDR mechanism is used to validate GNSS-based measurements.

The core technology used in our work is adapted from RouteNav, a wayfinding system for blind travelers in a transit hub developed by Peng et al. (2023). A notable addition to the localization module of RouteNav is the use of intermediate weights for particles falling on street surfaces. Our results show that this simple strategy substantially increases the accuracy of sidewalk assignment.

# Method

## Inertial Navigation via RoNIN

We use RoNIN (robust neural inertial navigation; Herath et al. 2020) for integration of inertial sensing data recorded by an iPhone. RoNIN is a state-of-the-art inertial navigation algorithm that uses deep learning to estimate the walker's velocity directly from inertial measurements. RoNIN operates at 200Hz, processing both IMU sensor data and 3D pose data, analyzing time-series windows of accelerometer and gyroscope readings to generate velocity estimates. A key technical feature of RoNIN is that it computes user velocity in a fixed reference frame rather than in the phone's reference frame. RoNIN accesses the phone's orientation (computed by integrating angular velocity from the gyroscope) then rotates all inertial data by the inverse of this rotation matrix, so that the inertial data is referred to a fixed reference frame. This coordinate transformation ensures that the user's walking velocity is measured with respect to a reference system that remains consistent regardless of how the user holds their device (e.g., if they hold it in their hand or in a pants' pocket), and of any device motion/rotation while walking (e.g. if the user checks the phone's screen before replacing the phone in their pocket). Of course, if the phone's orientation is incorrectly estimated (e.g., due to integration drift), the reference system and thus the user's estimated velocity will also be affected.

RoNIN processes 200-sample windows of accelerometer and gyroscope readings (representing one second of data) to generate velocity estimates. It was shown to be robust to different walking styles and phone placement. By integrating the velocity estimates from a known starting point and walking direction, the user's location is obtained. However, as with any PDR system, this location is subject to drift that, if unmitigated, can result in growing location error.

## Particle Filtering

Particle filtering is a probabilistic approach to position tracking that maintains $N$ hypotheses (particles) about the user's location. We used $N$=500 particles in our tests. Formally, each particle is a sample from the posterior distribution of a state given all prior observations. Following Peng et al. (2023), the state contained in each particle represents a location ($(x, y)$ coordinates) as well as a value for orientation drift angle θ and a weight $w$. Given the configuration of particles at time $t$, and a new velocity vector $v = (v_x, v_y)$ produced by RoNIN at time $t + \Delta t$, each particle is updated as follows. First, the vector $v$ is rotated by the rotation drift angle θ stored in the particle; the resulting vector, multiplied by $\Delta t$, is then added to the particle's location. Gaussian noise is added to the coordinates of the new particle's location, as well as to its orientation drift angle. The particle's weight $w$ is updated with a value that depends on the location of the particle in the map. Specifically, if the particle is at an "impenetrable" building, its weight is set to 0. If it is located on a freely traversable area (e.g., a sidewalk or a crosswalk), its weight is set to 1. If it is on a street surface, $w$ is set to an intermediate value, as explained later. All weights are then normalized such that their sum is equal to 1. At this point, all particles are resampled: $N$ new particles are sampled from the existing ones with replacement, where the probability of sampling a certain particle is equal to its weight. While typical implementations of particle filtering generally resample particles on a certain schedule, or based on the particles' statistics (e.g. their effective sample size, which measures the weight diversity across particles), we found that frequent resampling worked well for our application. Note that particles that end up in an "impenetrable" area, and thus have weight set to 0, are never resampled, so they effectively disappear. Particles on a street surface (outside of a crosswalk) are less likely to be resampled than particles on freely traversable areas. The orientation drift state component effectively tracks the drift in orientation of the velocity vectors produced by RoNIN, because only particles that have the correct orientation drift are propagated on freely traversable areas, rather than in impenetrable buildings. The user's location at each time is taken to be the weighted average of the particle locations.

## GNSS–Inertial Fusion

By combining information from GNSS and inertial localization, one can hope to obtain a good compromise: drift from dead-reckoning is kept under control by the zero-mean nature of GNSS measurements, while the potentially large GNSS noise is mitigated by the local consistency inherent in dead-reckoning integration. We use the particle filtering-based fusion strategy described by Peng et al. (2023), itself inspired by Ren & Ke (2010). This algorithm considers the radius of uncertainty of GNSS, which in iOS is provided by the `horizontalAccuracy` property of the `CLLocation` object. When the uncertainty radius is below a certain threshold, each particle's weight is multiplied by the value of a bivariate Gaussian density centered on the GNSS position and computed at the particle's location, where the the marginal standard deviations of this Gaussian density are proportional to the uncertainty radius. Eventually, the particle weights are normalized to unit sum. Intuitively, particles that are close to the GNSS location receive a higher weight, meaning that they contribute more to the final location (equal to the weighted sum of the particle locations), and are more likely to be resampled, thus skewing the particle distribution towards the GNSS location.

# Experiments

## Data collection

We recorded data from an iPhone carried by an experimenter walking along 6 different routes varying in length between 309 m and 644 m. These paths were chosen in areas of the City of San Francisco characterized by the presence of tall buildings. The length and main characteristics of these paths are summarized in Tab. 1, while the paths themselves are shown in Fig. 1. The first route was located in the Financial District and involved walking on sidewalks of large and small streets. The remaining routes were located in the South of Market area, near the Salesforce Tower (330 meters high, tallest building in San Francisco). Part of paths 2, 3 and 4 were in an alleyway between tall buildings. Path 5 and 6 were located inside the Salesforce Transit Center, in an open but covered area containing multiple bus stops (see Fig. 2, right). This location is particularly challenging for localization due to the lack of direct line of sight with GNSS satellites. For each route, we reported in Tab. 1 the number of street crossings on marked crosswalks, as well as the number of street crossings not on crosswalks.

**Table 1.** Characteristics of the paths traversed in our study.

| Path # | Length (m) | Street crossings (on crosswalks) | Street crossings (not on crosswalks) |
|---|---|---|---|
| 1 | 628 | 4 | 4 |
| 2 | 515 | 4 | 1 |
| 3 | 332 | 0 | 2 |
| 4 | 644 | 2 | 2 |
| 5 | 467 | 2 | 0 |
| 6 | 309 | 0 | 0 |

In order to implement the particle filter, we defined and labeled individual geosegments (represented in GeoJSON) indicating areas with (impenetrable) buildings and areas of street surfaces, except for marked crosswalks. While a walker's path is not expected to go through impenetrable buildings, walkers occasionally traverse a street, possibly even outside of a marked crosswalk ("jaywalking"). The remaining (unmarked) areas are considered to be "freely traversable"; they include sidewalks and other locations where it is safe to walk (including crosswalks). An example of geosegmentation is shown in Fig. 2 (left and center). This segmentation directly affects the weight of a particle at a certain location. A particle is assigned a weight of 0 when in an area labeled as an "impenetrable building"; a weight of 1 when in a "freely traversable" area; and a weight between 0 and 1 when in an area labeled as "street" (jaywalking). We experimented with three different weights (0, 0.4, 1) for particles in jaywalking areas. Remember that a particle's weight is used to estimate the user's location (as the weighted average of the particles), and represents the likelihood that the particle will be selected in the next round of resampling. Particles with weight of 0 do not contribute to the overall mean, and are not resampled (effectively, they disappear). Setting the jaywalking weight $w$ to 1 basically makes all of the surfaces unoccupied by buildings freely traversable. The intuition for using a lower weight than 1 for street surfaces stems from the consideration that pedestrians generally (but not always) cross streets on marked crosswalks. We choose a value of 0.4 as intermediate weight through preliminary tests in a smaller scale data set, created independently of the data collection presented here. For the purpose of sidewalk-based analysis, we also created polygonal segments representing individual sidewalks within each city block. All geosegments were created using Mapbox.

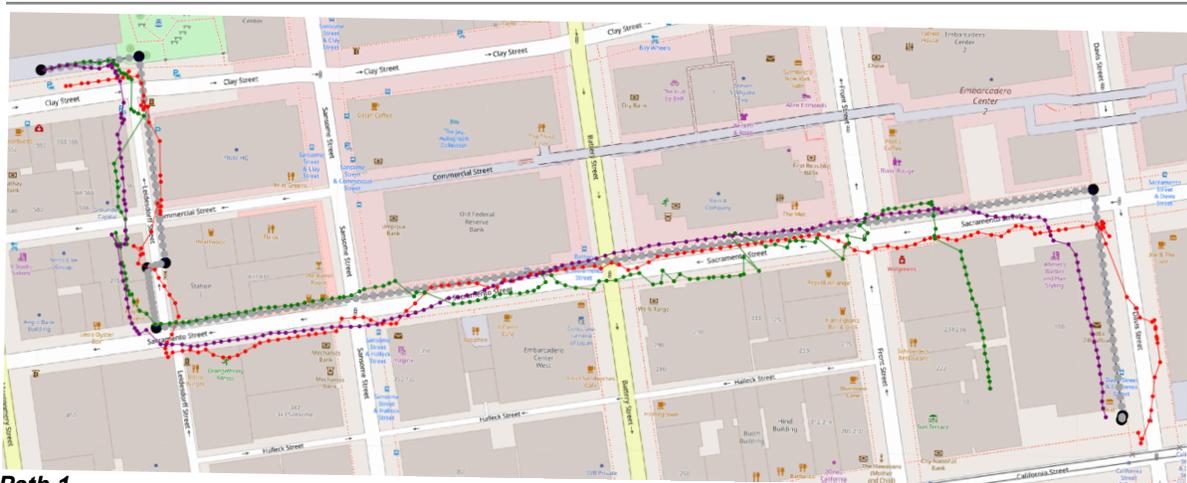

*Path 1*

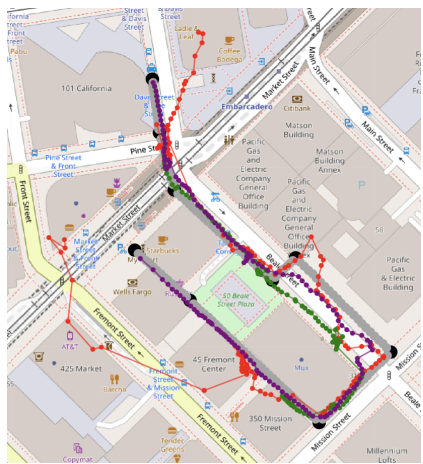

*Path 2*

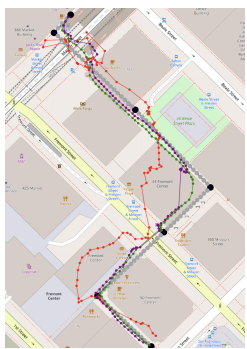

*Path 3*

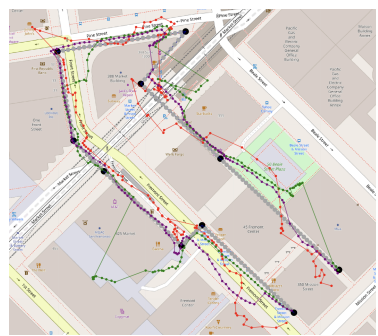

*Path 4*

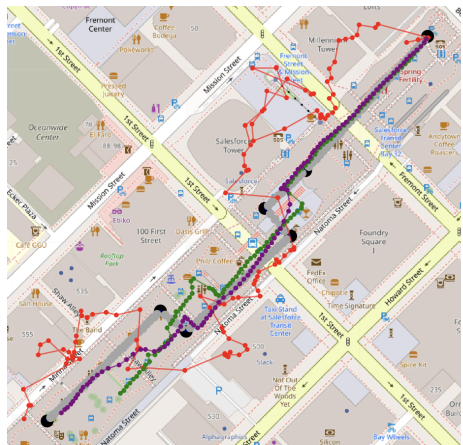

*Path 5*

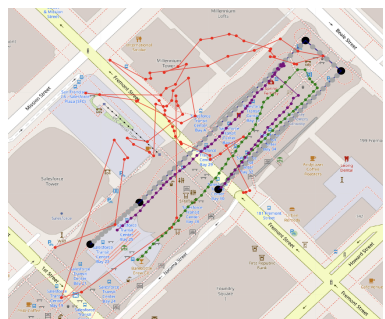

*Path 6*

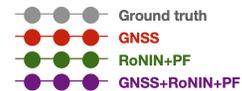

**Figure 1.** The paths traversed in our study. Waypoints are marked by dark circles. Ground truth locations at each footstep are shown by grey circles. GNSS, RoNIN+PF and GNSS+RoNIN+PF estimated locations at each footstep are shown by red, green, and purple circles, respectively.

For each path, an experimenter started at the starting location, then walked in a straight line at constant pace to the next designated waypoint (shown as black-filled circle in Fig. 1). The experimenter carried an iPhone 13 in their hand[1]. Once arrived at the waypoint, the user tapped a button on the smartphone's screen, and the current timestamp was recorded. From there, they started walking towards the next waypoint, and continued in this way until they reached the destination. GNSS and inertial data were recorded continuously, although particle filtering was only updated at each detected footstep, in order to minimize energy consumption. Fig. 1 shows the "ground-truth" (grey line) and estimated user location at each detected footstep (color-filled circles) based on three configurations:

1. *GNSS* (shown in red); this data is provided by iOS' `CoreLocation` framework and resampled at the times of detected footsteps.
2. Inertial-based localization based on the RoNIN algorithm, followed by particle filtering (*RoNIN+PF*, green).
3. Fused inertial-GNSS localization, using the algorithm described earlier (*GNSS+RoNIN+PF*, purple).

Note that for RoNIN+PF and GNSS+RoNIN+PF, the algorithm has access to the initial location and orientation of the user, which in our experiment was set manually. In practice, one could use a GNSS location measured with low uncertainty radius for initialization. For what concerns ground truth data (shown in grey in Fig. 1), we should emphasize that it is accurate at the waypoints, where the user manually tapped on the waypoint's location) on arrival; for the segments between consecutive waypoints, we made the assumption that the experimenter walked in a straight line and with approximately constant step length. An example of particle sets is shown for RoNIN+PF (green) and GNSS+RoNIN+PF (purple) in Fig. 3 (left). Note that in this case, the user was located on the sidewalk to the left of the visible building (between the red and its adjacent blue line), whereas the GNSS location was incorrectly set on the opposite sidewalk. The majority of particles for both systems were located on the correct sidewalk.

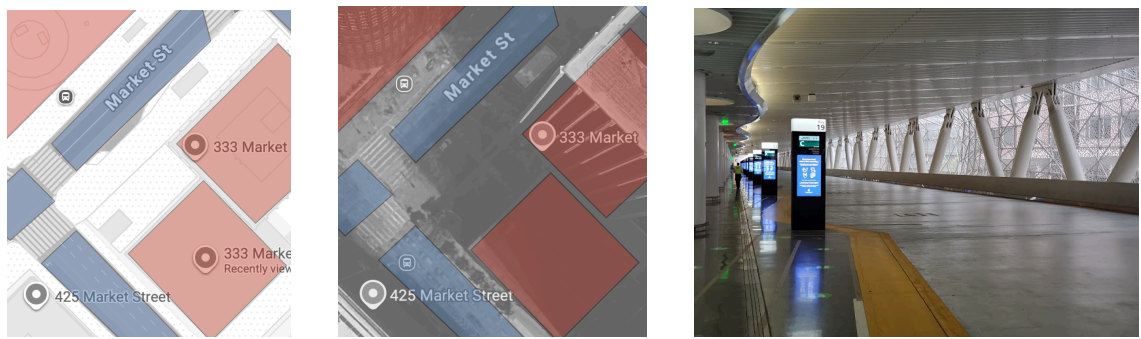

**Figure 2.** *Left and center*: Geosegments indicating impenetrable buildings (red) and street surfaces excluding crosswalks (blue). The same urban area is shown in two formats. *Right*: The environment where Paths 5 and 6 were located, which is a bus hub on the first floor of the Salesforce Transit Center.

## Error metrics

We considered three different metrics to evaluate the performance of the the considered localization/tracking methods:

1. **Correct Sidewalk Assignment.** For this metric, we first project the estimated user location at each footstep onto the closest sidewalk. Then, we check whether the projected estimated location is on the correct sidewalk segment (based on ground truth data). We then compute the proportion of correct sidewalk assignments to the total number of footsteps. In practice, this metric measures how often the user was localized in the correct city block, and on the correct side of the street.

2. **Euclidean Error**. This is the Euclidean distance between the estimated user location and the ground truth location at each footstep.

---

[1] For 4 of the paths (Paths 2, 3, 4, 6), the experimenter also carried an iPhone 13 in their pocket. Since the results after processing data from the two phones are almost identical, we only report results for the in-hand phone.

3. **Along- and Across-Street Errors**. The error vector at each footstep (joining the ground-truth location and the estimated location at each footstep) is decomposed in the along-street and the orthogonal across-street components, where the along-street direction is measured at the ground-truth location. The length of these two components is then measured individually.

We argue that Correct Sidewalk Assignment is the most relevant among the three considered metrics, as knowledge of the sidewalk segment where one is located is instrumental for providing guidance (e.g., to identify the next turn to take or whether one needs to cross the street). Euclidean Error gives a general indication of the goodness of tracking. The location error vector is known not to be uniformly oriented, but to depend on the direction of the street near the user's location. Along- and Across-Street Errors provide a characterization of the direction of this error relative to the street direction.

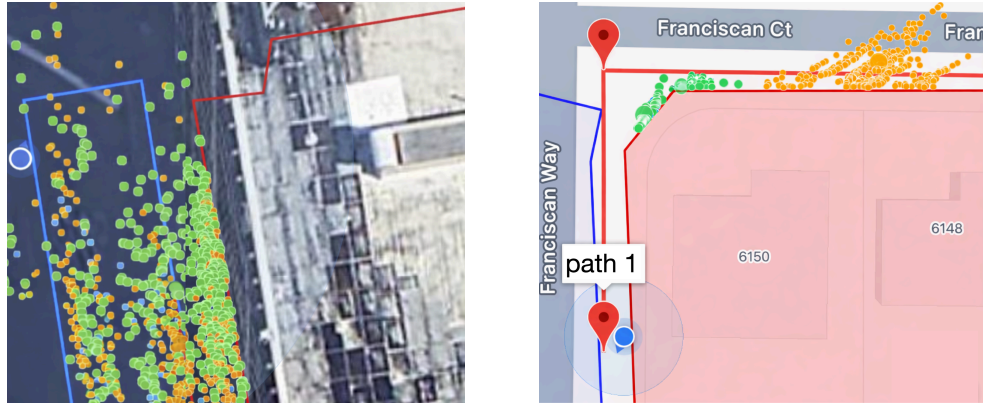

**Figure 3.** Visualization of particles in the particle filters in Path 1. Green = RoNIN+PF; Yellow = GNSS+RoNIN+PF. The large blue dot represents the GNSS location, along with the radius of uncertainty. The blue and red lines represent the borders or street surface and impenetrable building segments. The red markers in the right image represent waypoints along the path. Note that in the left image, the walker was located on the sidewalk next to the tall building, while the (incorrect) GNSS position is on the sidewalk across the street.

## Results

### Correct Sidewalk Assignment

Tab. 2 shows the Correct Sidewalk Assignment proportion for different values of the jaywalking weight $w$, averaged over all paths. (GNSS measurements do not undergo particle filtering, and thus are not affected by the choice of $w$.) Note that GNSS measurements are projected on the correct sidewalk segment less than half of the time. This proportion increases to 0.8 for the GNSS-inertial fusion algorithm when the proportion with jaywalking weight of 0.4. Since we consider this to be the most relevant metric for our purposes, we will use a jaywalking weight $w$ of 0.4 for further analyses unless otherwise noted.

**Table 2.** Correct Sidewalk Assignment proportions for the three methods, for different values of the "jaywalking weight" $w$. Note that $w$ does not affect measurements from GNSS, as they do not undergo particle filtering.

|         | GNSS | RoNIN+PF | GNSS+RoNIN+PF |
|---------|------|----------|---------------|
| $w = 0$   | 0.47 | 0.21     | 0.30          |
| $w = 0.4$ | 0.47 | 0.71     | **0.80**      |
| $w = 1$   | 0.47 | 0.41     | 0.72          |

## Euclidean Error

Fig. 4 shows the cumulative distribution function (CDF) of the Euclidean error for all measured localization samples using GNSS, inertial localization (RoNIN+PF), and fused localization (GNSS+RoNIN+PF). Mean, median, and 90th percentile values are reported in Tab.3.

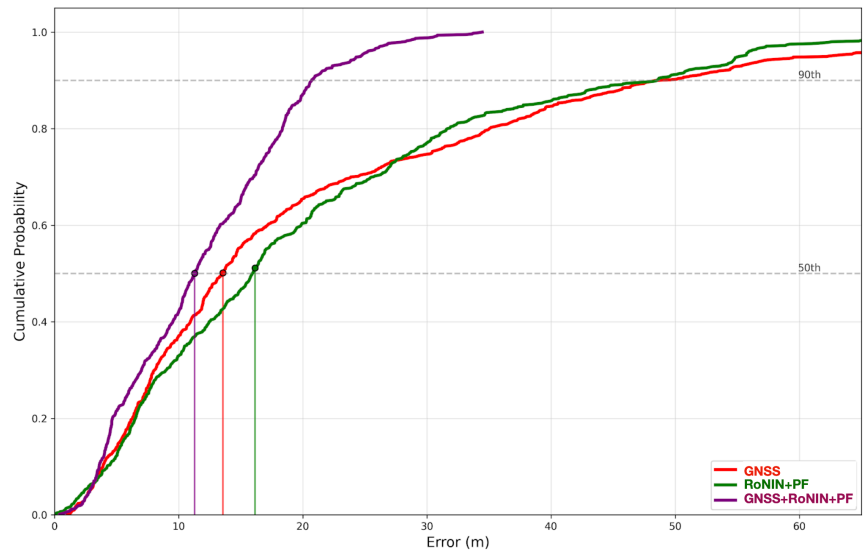

**Figure 4.** Cumulative distribution function of Euclidean error for all measurements for the three methods considered.

GNSS+RoNIN+PF is seen to perform better than the other two methods. While the median error with GNSS+RoNIN+PF is only 17% (2.3 m) smaller than using GNSS alone, large errors are substantially less likely to be found using GNSS+RoNIN+PF than with either GNSS or RoNIN+PF (e.g., the 90th percentile error is reduced by 57%, or more than 25 m, in both cases). For example, the CDF plot shows that errors of 30 or more meters were almost never found using the GNSS+RoNIN+PF, yet they accounted for more than 20% of the measurements with the other methods. Interestingly, a higher value of the jaywalking weight $w$ was shown to give conflicting results. For $w = 1$ (meaning that particles in jaywalking areas are not penalized), the median Euclidean error increases by 2.8 m to 18.9 m for RoNIN+PF, while it reduces by 1.7 m to 9.6 m for GNSS+RoNIN+PF. On the contrary, reducing the weight $w$ to 0 (maximum penalty; particles not allowed in a street surface except for crosswalks) leads to substantially higher errors overall (median error of 52.0 m RoNIN+PF and of 43.8 m for GNSS+RoNIN+PF).

**Table 3.** Statistics of Euclidean Error for the three methods considered.

|  | GNSS | RoNIN+PF | GNSS+RoNIN+PF |
|---:|:---:|:---:|:---:|
| **Mean (m)** | 20.9 | 23.7 | **11.7** |
| **Median (m)** | 13.6 | 16.1 | **11.3** |
| **90th percentile (m)** | 48.2 | 47.4 | **20.5** |

Fig. 5 shows the Euclidean error averaged within individual paths. Note that the quality of GNSS localization varied substantially across paths. For Path 1, the average GNSS error was about 10 meters, which was lower than obtained with the other methods. For all other paths, GNSS+RoNIN+PF gave substantially lower errors. In one case (Path 3), inertial-based localization (RoNIN+PF) gave a better result than fusion with GNSS (GNSS+RoNIN+PF), likely due to the poor quality of GNSS in that path. For Paths 5 and 6, which were located in a covered area (Fig. 2, right), GNSS was highly unreliable (see also Fig. 1) Despite that, fusing information from GNSS with inertial localization (GNSS+RoNIN+PF) gave the best results. For example, in Path 5, GNSS had an average error of almost 30 meters, worse than by inertial sensing alone (RoNIN+GNSS; mean Euclidean error = 25 m). By fusing together information from GNSS and inertial sensing (GNSS+RoNIN+PF), the average Euclidean error was reduced to 11 m.

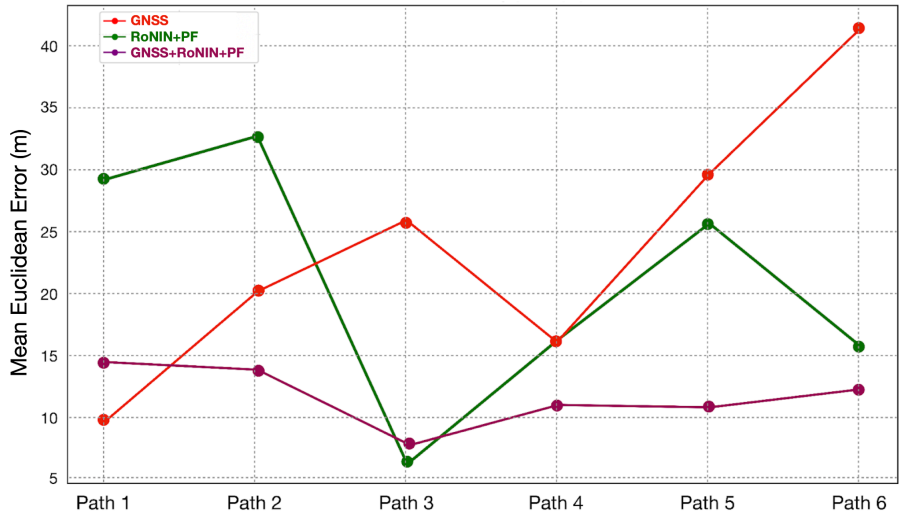

**Figure 5.** Euclidean error averaged over each path for each considered method.

## Along- and Across-Street Error

Tab. 4 shows the Along- and Across-Street median error for the different methods (with jaywalking weight $w$ set to 0.4).

**Table 4.** Along- and across-street median errors for the three considered methods.

|  | GNSS | RoNIN+PF | GNSS+RoNIN+PF |
|---|---|---|---|
| **Along-street median error (m)** | **6.2** | 11.3 | 9.0 |
| **Across-street median error (m)** | 8.1 | 4.7 | **3.1** |

Consistent with literature (Wang et al., 2015), GNSS localization was more accurate along-street (median error: 6.2 m) than across-street (median error: 8.1 m). Interestingly, while both other methods gave better across-street results than GNSS, they both had larger along-street errors. In particular, the median along-street error for GNSS+RoNIN+PF was 9.0 m, whereas its median across-street error was 3.1 m. (Note that, as in the case of Euclidean Error, GNSS+RoNIN+PF had lower likelihood of producing large errors. E.g., the 90-th percentile of along-street errors was 19 m for GNSS+RoNIN+PF, vs. 34 m for GNSS.) This behavior can be justified in part by observing that, when a path includes a turn around a corner, particles often accumulate at the corner before turning, generating a "lag". This is especially noticeable when the tracker undershoots the walker's position (i.e., particles lag behind the true location). For example, consider the situation shown in Fig. 3 (right), where green and yellow dots are particles for RoNIN+PF and GNSS+RoNIN+PF, respectively, and the large blue dot is the GNSS location. Intuitively, GNSS "pulls" all particles towards itself, but the particles are impeded by the presence of an impenetrable building. In general, the noise applied to the particle ensures that eventually a portion of them negotiate the corner and quickly move towards the GNSS location, but this "accumulation" is responsible for a substantial (if temporary) lag increase, which contributes to the along-street error.

## Conclusions

We have presented an empirical analysis of the potential benefits of inertial-based pedestrian localization in urban areas characterized by poor GNSS signal reception, due to the presence of tall buildings (urban canyons). We have considered two algorithms: the first one (RoNIN+PF) measures the user's location using a machine learning dead-reckoning algorithm (RoNIN), followed by a particle filter that leverages geographical information in the form of impenetrable buildings, street surface areas that are unlikely to be crossed, and freely traversable areas. This method does not use GNSS information at all, but it requires knowledge of starting location and orientation. Street crossing outside of crosswalks (jaywalking) is tolerated by the particle filter, but with a weight that reflects the fact that jaywalking is a sporadic event. The second algorithm (GNSS+RoNIN+PF) fuses information from GNSS, thus helping to reduce drift and limiting the chance of catastrophic errors (e.g. consequent to particles dispersing over multiple directions for extended periods of time). We tested these algorithms over long paths in realistic and challenging urban scenarios, including a transportation hub with highly unreliable GNSS localization.

Under the Correct Sidewalk Assignment metric, which only identifies the sidewalk segment closest to the user location as provided by the algorithms (thus implementing a simple form of map-matching), GNSS+RoNIN+PF was shown to identify the correct sidewalk 80% of the times, vs. 71% for RoNIN+PF and 47% for GNSS alone. Our analysis highlighted the importance of choosing an appropriate value for the "jaywalking weight", which basically quantifies the notion that walkers are expected to generally walk on sidewalks or crosswalks, though may occasionally cross a street outside of a crosswalk. GNSS+RoNIN+PF also produced the smallest Euclidean error. In particular, it was shown to generate a much lower proportion of measurements with large errors than the other methods. When looking at along- and across-street errors, results are more mixed. The lowest along-street error was measured with GNSS (6.2 m median error, vs. 9.0 m for GNSS+RoNIN+PF), whereas GNSS+RoNIN+PF gave the lowest across-street error (3.1 m median error, vs. 8.1 m for GNSS).

Leveraging inertial sensing for situations when GNSS accuracy is poor is a well established technique. This work has shown that the use of machine learning-based PDR techniques (such as RoNIN), combined with particle filtering, can be extremely beneficial. While this method has been used before in the context of indoor wayfinding (see e.g. Tsai et al., 2024), we have shown that this same technique is feasible outdoors (possibly complementing other methods that process GNSS data for improved location accuracy), with simple adjustments to account for the likelihood of walking in specific areas. Critically, for particle filtering to be effective, a labeled map of impenetrable buildings and street surfaces is available. While labeled geosegments were obtained manually in this work, in most cases a similar segmentation can be obtained automatically from analysis of GIS data. For example, OpenStreetMap marks, for most densely inhabited areas, the location and extent of buildings, as well as the location of surface streets, sidewalks and crosswalks.